\documentclass{article}
\usepackage{spconf,amsmath,graphicx}
\usepackage{caption}
\usepackage{ctable}
\usepackage{multirow}
\usepackage{amssymb}

\title{Self-supervised Contrastive Learning\\for Cross-domain Hyperspectral Image Representation}
\name{Hyungtae Lee~~~~~~~~~~~~~~~~~~~~~~~~~~~Heesung Kwon\thanks{© 2022 IEEE. Personal use of this material is permitted. Permission from IEEE must be obtained for all other uses, in any current or future media, including reprinting/republishing this material for advertising or promotional purposes, creating new collective works, for resale or redistribution to servers or lists, or reuse of any copyrighted component of this work in other works.}}
\address{DEVCOM Army Research Laboratory (ARL)}
\begin{document}
%
\maketitle
\begin{abstract}
Recently, self-supervised learning has attracted attention due to its remarkable ability to acquire meaningful representations for classification tasks without using semantic labels. This paper introduces a self-supervised learning framework suitable for hyperspectral images that are inherently challenging to annotate. The proposed framework architecture leverages \emph{cross-domain CNN}~\cite{HLeeIGARSS2018}, allowing for learning representations from different hyperspectral images with varying spectral characteristics and no pixel-level annotation. In the framework, cross-domain representations are learned via contrastive learning where neighboring spectral vectors in the same image are clustered together in a common representation space encompassing multiple hyperspectral images. In contrast, spectral vectors in different hyperspectral images are separated into distinct clusters in the space. To verify that the learned representation through contrastive learning is effectively transferred into a downstream task, we perform a classification task on hyperspectral images. The experimental results demonstrate the advantage of the proposed self-supervised representation over models trained from scratch or other transfer learning methods.

\end{abstract}
\section{Introduction}
\label{sec:intro}

Self-supervised learning is to learn representations on unlabeled data in a supervised manner. Recently, many downstream tasks, such as image classification, started using a self-supervised approach due to its remarkable capability of representing underlying characteristics of data without relying on data labels. In particular, self-supervised learning is suitable for tasks that require large-scale image/video data whose annotations are quite laborious and expensive (e.g. 1B images~\cite{KHeCVPR2020}, 240k videos~\cite{CFeichtenhoferCVPR2021}) or tasks that entail inherently challenging labeling. Hyperspectral image classification, often used in remote sensing applications, falls into the second category, where most pixels (i.e., spectral vectors) remain unlabeled primarily because manual pixel-level annotation for remotely sensed data is laborious and often erroneous. This paper presents a self-supervised method that initially learns data representations from unlabeled hyperspectral images then transfer the learned representation to downstream tasks, such as hyperspectral image classification.

Conventional self-supervised learning starts with designing pseudo-labels that automatically generate samples that are assigned to multiple groups. Then, it adopts contrastive learning~\cite{RHadsellCVPR2006} to learn data representation in such a way that samples from the same group are closely clustered together, and samples of different groups are separated apart in the representation space. Pseudo-labels can be determined based on the type of given tasks. For image classification tasks~\cite{KHeCVPR2020,TChenICML2020,MCaronNeurIPS2020,JBGrillNeurIPS2020,XChenCVPR2021,XChenICCV2021}, samples are randomly cropped at different locations of the same image, which are then assigned to the same group. Similarly, different temporal segments of the same video in video-based applications~\cite{CFeichtenhoferCVPR2021} and the images containing the same view from different sensors in multi-view applications~\cite{YTingECCV2020} are assigned to the same group.

\begin{figure}[t]
    \captionsetup{font=small}
    \centering
    \includegraphics[trim=5mm 5mm 5mm 5mm,clip,width=0.85\linewidth]{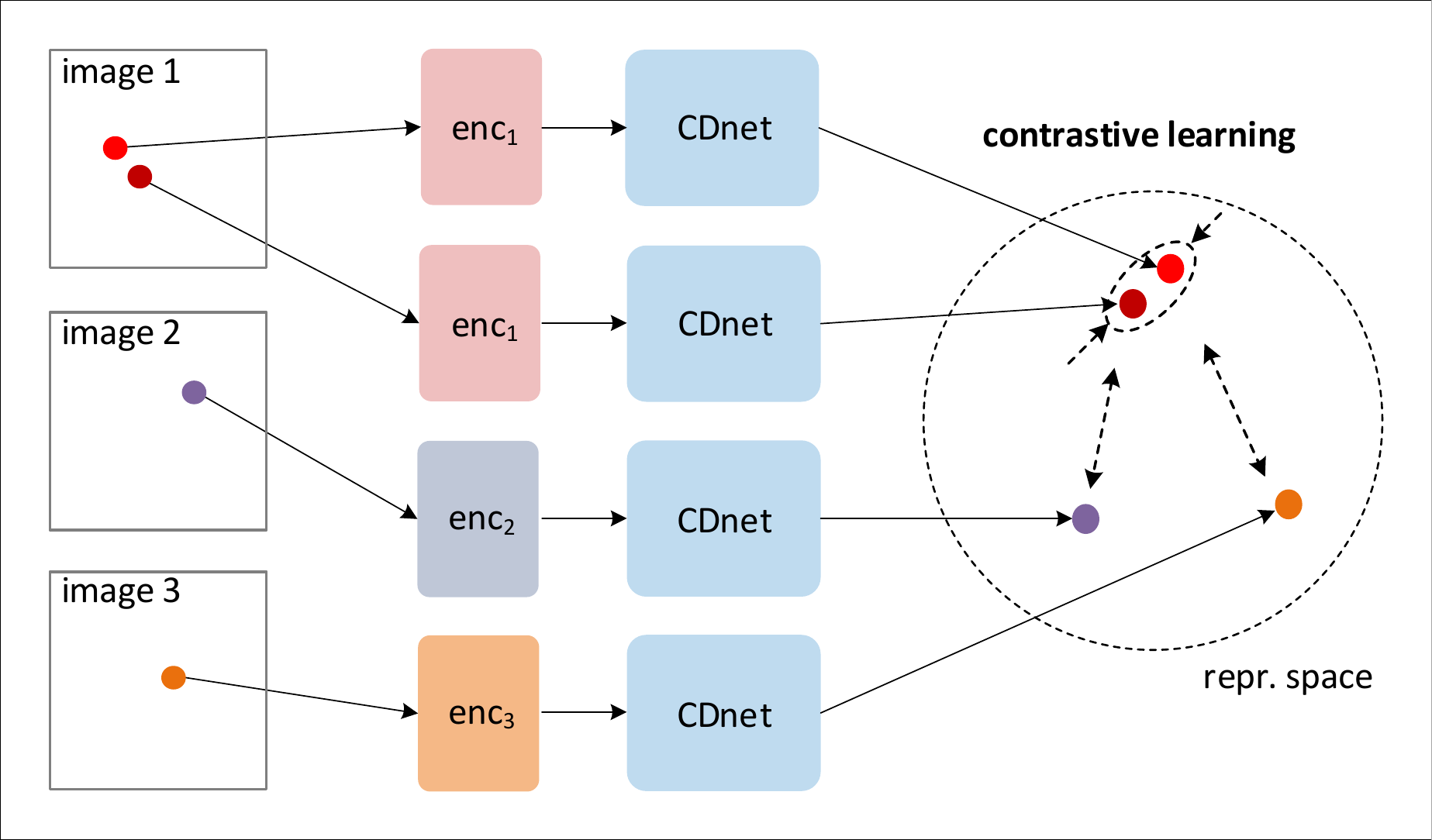}
    \vspace{-0.3cm}
    \caption{{\bf Contrastive learning for cross-domain network.} enc$_i$ and CDnet are a domain-specific feature encoder for the $i^\text{th}$ spectral domain and a shared cross-domain network across all the domains, respectively.}
    \label{fig:concept}
\end{figure}

In this paper, we aim to acquire a \emph{cross-domain hyperspectral representation} learned from multiple hyperspectral images with a range of different spectral characteristics using self-supervised learning\footnote{In this paper, we refer to a single hyperspectral image as a \emph{hyperspectral domain} because each image contains different spectral properties is associated with a very different classification task.}. Accordingly, we design pseudo-labels suitable for building the cross-domain representation across multiple hyperspectral domains. Specifically, we randomly select a single $p\times p$ region from each image for every training iteration. When $p$ is a small value, we consider pixels in the same region to belong to the same group since neighboring pixels in remote sensing images are highly likely from the same material or terrain. We also consider pixels from different images as coming from other groups. A simple example of this grouping and how contrastive learning is applied to grouping is shown in Figure~\ref{fig:concept}.

Our method of acquiring cross-domain hyperspectral representation, has been inspired in part by a \emph{cross-domain CNN framework} (CD-CNN)~\cite{HLeeIGARSS2018}, which can be applied on multiple hyperspectral images with different spectral properties (Figure~\ref{fig:concept}). CD-CNN architecture consists of a domain-specific feature encoder, a shared cross-domain network, and a domain-specific classification layer. The domain-specific feature encoder and the domain-specific classification layer are each designed to deal with different spectral bands and different classification categories, respectively. Since the hyperspectral image representation is learned in a self-supervised manner that does not depend on given class labels, domain-specific classification layers are not used.

Since the transferability of the representation into a downstream task is crucial to self-supervised learning, we tested our model for hyperspectral image classification through transfer learning. We selected six different hyperspectral images, among which the Indian pines image was used for a downstream task and the remaining images were used to learn the cross-domain hyperspectral image representation. Experiments on the image classification task show that our representation presents significantly improved performance over other transfer learning methods or models trained from scratch. This demonstrates that the proposed self-supervised learning framework successfully learns the transferable representation given many unlabeled samples.

\section{Method}
\label{sec:method}

\subsection{Self-supervised Contrastive Learning}
\label{ssec:self-supervised}

\noindent{\bf Pseudo-labeling for a self-supervised task.} To design pseudo-labels suitable for building cross-domain hyperspectral representations, we use fundamental properties of remotely sensed hyperspectral images; the spectral pixel vectors belonging to the same regions have similar spectral properties. On the other hand, since different hyperspectral images are taken from different areas with potentially different hyperspectral sensors, it is highly likely that pixel vectors in different images do not belong to the same category. We verify that the labeled categories do not overlap each other in all the hyperspectral images used in the experiment.

Accordingly, for every training iteration, a single $p\times p$ region is randomly selected in each image, and the pixel vectors in the region are considered samples from the same group. We do not select more than one region from one image as these regions may belong to the same category. Also, the pixels from different images are considered samples of different groups.\smallskip

\noindent{\bf Contrastive learning.} Conventional contrastive learning methods are performed in a setting where there are query samples and corresponding sets of one positive sample and multiple negative samples for each query. Unlike the conventional methods, we have to consider the case where there are multiple positive samples as $p\times p$ pixels fall into the same category in our problem.

In training samples, each sample can be considered a query ($q$), which is compared to the remaining samples ($k$) in the representation space. A contrastive loss measures the similarity of $q$ and $k$ and outputs lower values for more similar pairs. In general, a InfoNCE loss~\cite{AOordArXiv2018} and its variant form are used as the contrastive loss. We also use a variant of InfoNCE, modified to account for multi-label cases where multiple positive samples are allowed, as below:
\begin{equation}
    \mathcal{L}_q = \sum_{k_+ \in \mathcal{R}(q)}{\left(-\text{log}{\frac{\text{exp}(q\cdot k_+ / \tau)}{\sum_{i=0}^K \text{exp}(q\cdot k_i / \tau)}}\right)},
\end{equation}
where $\mathcal{R}(q)$ is a set consisting of samples ($k_+$) belonging to the region to which the query sample belongs. $\tau$ is a temperature hyper-parameter. The expression in parentheses is the single-label InfoNCE loss function that allows only one positive sample for the query. This modification was derived from a multi-label cross-entropy loss defined as the sum of the single-label cross-entropy loss, where the InfoNCE loss has the form of a single-label cross-entropy loss.

\subsection{Implementation Details}
\label{ssec:training_and_inference}

\begin{table}[t]
\captionsetup{font=small}
\setlength{\tabcolsep}{9.7pt}
\renewcommand{\arraystretch}{0.9}
\centering
{\small
\begin{tabular}{l|l|l}
layer name & \multicolumn{1}{c|}{\cite{HLeeTIP2017}} & \multicolumn{1}{c}{our backbone} \\\specialrule{.15em}{.05em}{.05em}
\multirow{4}{*}{C1} & 1$\times$1, 128 & \multirow{4}{*}{5$\times$5, 128, pad 2}\\
& 3$\times$3, 128, pad 2$^1$ & \\
& 5$\times$5, 128, pad 4$^2$ & \\\cline{2-2}
& channel-wise concat. & \\\hline
C2 & 1$\times$1, 128 & \multicolumn{1}{c}{$\cdot$} \\\hline
\multirow{3}{*}{Rx} & \multirow{3}{*}{$\begin{bmatrix}1\times 1, 128\\1\times 1, 128\end{bmatrix}$$\times$2} & \multirow{3}{*}{$\begin{bmatrix}1\times 1, 128\\1\times 1, 128\end{bmatrix}$$\times n$} \\
&&\\
&&\\\hline
C3 & 1$\times$1, 128 & \multicolumn{1}{c}{$\cdot$} \\\hline
C4 & 1$\times$1, 128 & \multicolumn{1}{c}{$\cdot$} \\\hline
C5 & 1$\times$1, $C$ & 1$\times$1, $C$ \\\specialrule{.15em}{.05em}{.05em}
FLOPs$^3$ & 43.9$\times 10^9$ & 33.7$\times 10^9$ ($n=5$) \\
\multicolumn{3}{l}{$^1$ 3$\times$3 max pooling is applied to the output} \\ 
\multicolumn{3}{l}{$^2$ 5$\times$5 max pooling is applied to the output} \\ 
\multicolumn{3}{l}{$^3$ FLOPs is calculated as the equation given in \cite{HLeeDCS2019}.} \\ 
\end{tabular}
}
\vspace{-0.3cm}
\caption{{\bf The proposed backbone architecture.} Each layer is indicated by the filter size, number of filters, and padding size (if applicable). $C$ represents the number of categories in a hyperspectral image. FLOPs (floating-point operations, multiply \& addition) are calculated assuming that the Indian pines dataset is used.}
\label{tab:backbone_architecture}
\end{table}

\noindent{\bf Cross-domain CNN architecture.} Our architecture (Figure~\ref{fig:concept}) has the ability to take inputs from any type of a hyperspectral domain. This capability was acquired by designing our model inspired by Cross-domain CNN (CD-CNN)~\cite{HLeeIGARSS2018} that enables model training on multiple hyperspectral domains simultaneously. CD-CNN consists of a domain-specific feature encoder, a shared cross-domain network, and a domain-specific classification layer, of which the first two modules were adopted as the feature encoder and the cross-domain network in our model, respectively.

The backbone of CD-CNN uses a 9-layer hyperspectral image classification CNN~\cite{HLeeTIP2017,HLeeIGARSS2016,HLeeJSTARS2021} consisting of the first two multi-scale layers, two two-layer residual modules, and three final layers. The first layer use filters with various sizes (i.e., 1$\times$1, 3$\times$3, and 5$\times$5), while the other layers only use 1$\times$1 filters. This architecture was modified to achieve optimal accuracy in our task, resulting in an initial layer with a 5$\times$5 filter replacing the multi-scale layers, followed by $n$ two-layer residual modules ($n=5$ in our experiment). Then, we use the initial layer as a domain-specific feature encoder and the remaining layers as a shared cross-domain network. These modifications are shown in Table~\ref{tab:backbone_architecture}.\smallskip

\noindent{\bf Transferring to a downstream task.} The goal of self-supervised learning is to learn a transferable representation. With this goal, we follow the pretrain-finetune strategy in \cite{HLeeIGARSS2019} designed for cross-domain hyperspectral image classification in transferring a learned representation to the downstream task. Specifically, only a middle cross-domain shared network is fine-tuned for a downstream task because the domain-specific feature encoders of the representation cannot be transferred due to the difference in spectral properties. The domain-specific networks (i.e., the first and last layers) are trained from scratch with a 10$\times$ learning rate than that used to train the transferred layers. Adopting different learning rate was crucial to enhancing accuracy from this transfer. We use cross-entropy loss in training the downstream model.\smallskip

\noindent{\bf Training.} We use the SGD optimizer with the momentum of 0.9, gamma of 0.1, and weight decay of 0.005. The pre-trained model is trained with an initial learning rate of 0.03 for 200 iterations. The learning rate was reduced by 1/10 at 120 and 160 iterations. The downstream model is trained with an initial learning rate of 0.03 for 100 iterations. The learning rate was reduced by 1/10 at 60 and 80 iterations. All the layers that are not inherited from the pre-trained model are initialized according to the Gaussian distribution with zero-mean and standard deviation of 0.001. To provide a richer set of training samples, we use eight-fold data augmentation by mirroring each sample across the vertical, horizontal, and two diagonal axes~\cite{HLeeTIP2017}.

\section{Experiments}
\label{sec:experiments}

\subsection{Setting}
\label{ssec:setting}

\begin{table}[t]
\captionsetup{font=small}
\setlength{\tabcolsep}{8pt}
\renewcommand{\arraystretch}{0.9}
\centering
{\small
\begin{tabular}{l|c|c|c|c}
image & sensor & bands & \# data & \# labeled \\\specialrule{.15em}{.05em}{.05em}
Indian pines & \multirow{3}{*}{AVIRIS} & 200 & ~~21,025 & ~~~~9,545 \\\cline{1-1}\cline{3-5}
Salinas && 204 & 111,104 & ~~54,129 \\\cline{1-1}\cline{3-5}
KSC && 176 & 314,368 & ~~~~5,211 \\\hline
Pavia & \multirow{2}{*}{ROSIS} & 102 & 783,640 & 148,152 \\\cline{1-1}\cline{3-5}
Pavia Univ. & & 103 & 207,400 & ~~42,776 \\\hline
Botswana & Hyperion & 145 & 377,856 & ~~~~3,248 \\
\end{tabular}
}
\vspace{-0.3cm}
\caption{{\bf Hyperspectral images.} Images acquired by the same type of a sensor can have different spectral bands because some spectral bands are removed according to the task unique to each image.}
\label{tab:dataset}
\end{table}

\noindent{\bf Dataset.} We use six hyperspectral images shown in Table~\ref{tab:dataset}. The Indian pines is sequestered for the downstream task and the remaining images are used to learn cross-domain representations. Reported in the following experiments are accuracy values for the downstream Indian pines image. Unless specified, 200 samples are randomly selected as training samples and the remaining samples are used to evaluate the model for downstream tasks.\smallskip

\noindent{\bf Metric.} We use OA (overall accuracy) and AA (average accuracy) as evaluation metrics, which are widely used in the literature on hyperspectral image classification. We reported average accuracy for 5 runs as different training/test splits for each run affect accuracy.\smallskip

\noindent{\bf Default specification.} Unless specified, we use $p$ = 6 as the default specification for pseudo-labeling. In addition, unless specified, we use $n=5$, which is the number of a residual module in the backbone architecture. We set $\tau$ to 0.7 that defines the contrastive loss to 0.07.

\begin{table}[t]
\includegraphics[trim=15mm 85mm 15mm 95mm,clip,width=.85\linewidth]{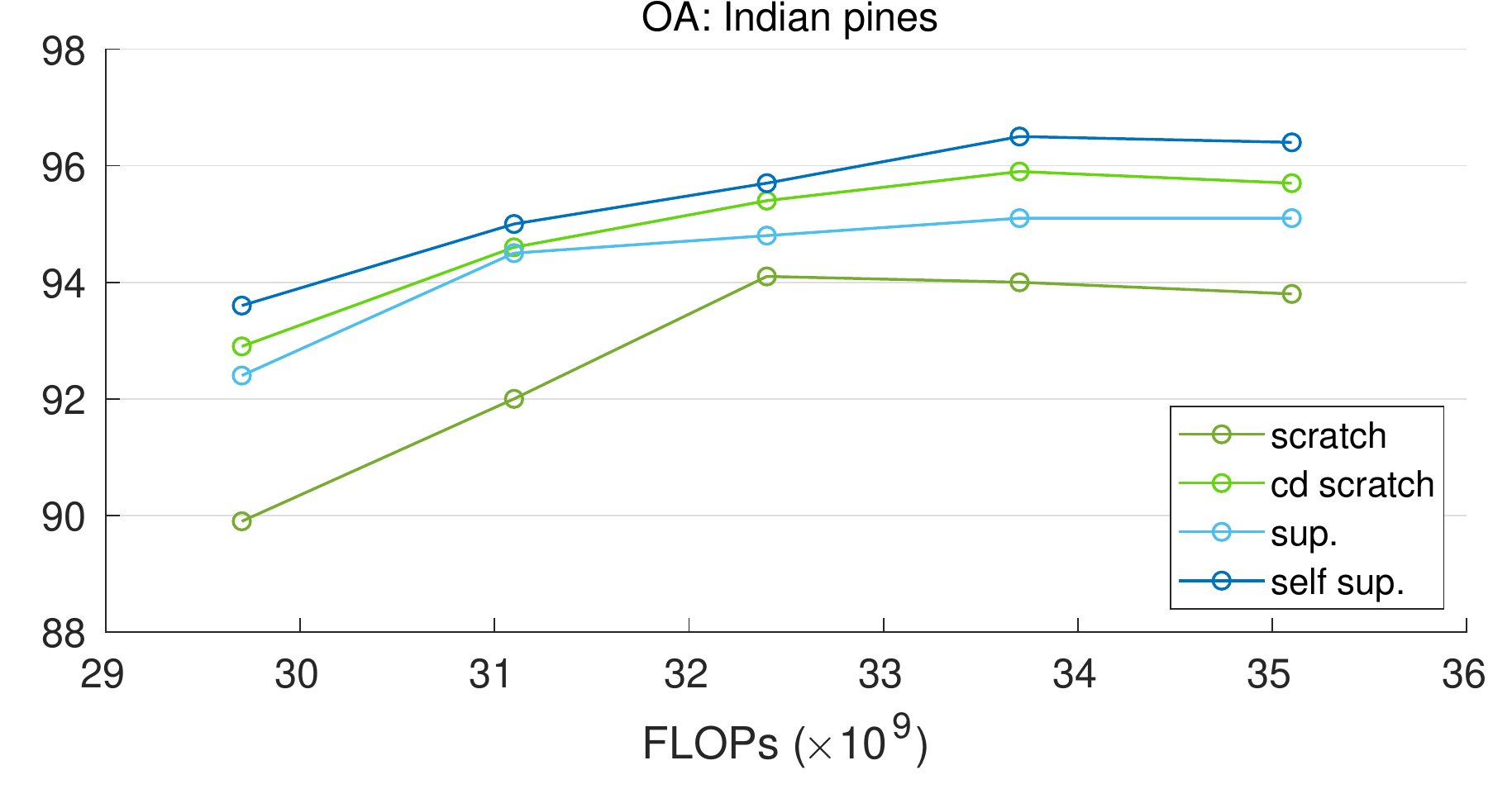}\\
\vspace{-0.5cm}
\captionsetup{font=small}
\setlength{\tabcolsep}{8.0pt}
\renewcommand{\arraystretch}{0.9}
\centering
{\small
\begin{tabular}{l|c|cc}
pretrain & \# of res. module & OA & AA \\\specialrule{.15em}{.05em}{.05em}
\textcolor{gray}{scratch} & 4 & 94.1~~~~~~~~~~~ & 92.3~~~~~~~~~~~ \\\hline
cd. scratch & 5 & 95.9 \textcolor{teal}{(+1.8)} & 93.1 \textcolor{teal}{(+0.8)} \\
sup. & 5 & 95.1 \textcolor{teal}{(+1.0)} & 92.9 \textcolor{teal}{(+0.6)} \\
{\bf self-sup.} & 5 & {\bf 96.4} {\bf \textcolor{teal}{(+2.3)}} & {\bf 94.3} {\bf \textcolor{teal}{(+2.0)}} \\
\end{tabular}
}
\vspace{-0.2cm}
\caption{{\bf Comparisons of transfer learning methods.} The top figure presents FLOPs vs. OA for the comparison methods including training-from-scratch model and three transfer learning models. The optimal accuracy of these models is shown in the bottom table. In the parentheses, gaps to ``scratch'' are shown.}
\label{tab:different_model_depth}
\end{table}

\subsection{Transfer Learning}
\label{ssec:transfer_learning}

We evaluate the hyperspectral representation trained by the proposed self-supervised learning framework in a downstream task to validate its transfer capability. We compare our model (``self-sup.'') with one model not relying on any transferable features (``scratch'') and two models using different forms of transfer learning. One transfer learning model (``cd scratch'') is trained on all six hyperspectral images together with a cross-domain CNN framework. Another transfer learning model (``sup.'') is acquired by taking a pretrain-finetune strategy, as done for our model. Here the pre-trained model is trained using labeled samples only in a supervised manner. All transfer learning methods used features transferred from five hyperspectral images not used as the downstream image. In addition, our model is the only method that uses unlabeled samples in training a pre-trained model.

We compare the accuracy of these models by varying the model depths, as shown in Table~\ref{tab:different_model_depth}. The reason we compared models of different depths is that with transfer learning, using more data can mitigate overfitting, which leads to higher accuracy by adopting deeper models. As expected, the three transfer learning methods can have deeper models leading to improved OA (5 vs. 4 in number of residual modules). 

Our method using self-supervised learning outperforms all the methods. The improved accuracy is mainly attributed to  \emph{the proposed self-supervised learning optimally capturing the transferable representation from large amounts of unlabeled samples.} In addition, it is noteworthy that ``cd scratch'' yields the better OA than ``sup.'', where both methods use only labeled samples for training. This could be interpreted as joint training is more effective than using a pretrain-finetune strategy regarding the transferable capability.

\subsection{Ablation Experiments}
\label{ssec:ablation}

\begin{table}[t]
\captionsetup{font=small}
\setlength{\tabcolsep}{15.5pt}
\renewcommand{\arraystretch}{0.9}
\centering
{\small
\begin{tabular}{l|c|cc}
pretrain & $p$ & OA & AA \\\specialrule{.15em}{.05em}{.05em}
\textcolor{gray}{scratch} & $\cdot$ & 94.1~~~~~~~~~~~ & 92.3~~~~~~~~~~~ \\\hline
\multirow{6}{*}{self-sup.} & 2 & 95.8 \textcolor{teal}{(+1.7)} & 94.3 \textcolor{teal}{(+2.0)} \\
& 3 & 95.9 \textcolor{teal}{(+1.8)} & 94.1 \textcolor{teal}{(+1.8)} \\
& 4 & 96.6 \textcolor{teal}{(+2.5)} & 93.8 \textcolor{teal}{(+1.5)} \\
& 5 & 96.4 \textcolor{teal}{(+2.3)} & 94.3 \textcolor{teal}{(+2.0)} \\
& 6 & 95.1 \textcolor{teal}{(+1.0)} & 93.3 \textcolor{teal}{(+1.0)} \\
& 7 & 94.8 \textcolor{teal}{(+0.7)} & 93.0 \textcolor{teal}{(+0.7)} \\
\end{tabular}
}
\vspace{-0.3cm}
\caption{{\bf Effect of region cropping size on pseudo-labeling.}}
\label{tab:crop_resion_size}
\end{table}

\noindent{\bf $p$ in pseudo-label.} Our pseudo-labeling strategy assigns all samples from the randomly selected $p\times p$ region per image to one group. In this strategy, it is important to maintain a level of the spectral homogeneity of each category within the group, so the region size should be carefully determined. As the region grows larger, it becomes more difficult to maintain the spectral homogeneity, which negatively affects the quality of the representation. On the other hand, we can also expect benefits from using a larger number of samples at the same time. To investigate the relationship between accuracy and region size, denoted by $p$, we compared the accuracy of pre-trained models according to various pseudo-labeling strategies using different $p$ in Table~\ref{tab:crop_resion_size}. From the comparison, we observed that when a small number of $p < 5$ was used, there was no significant differences in accuracy, but the accuracy was lowered when $p\geq6$. These observations indicate the benefit of using more samples by taking a large region from each image for pseudo-labeling affects accuracy less than the adverse effect due to incorrect labeling.\smallskip

\begin{table}[t]
\captionsetup{font=small}
\setlength{\tabcolsep}{5.0pt}
\renewcommand{\arraystretch}{0.9}
\centering
{\small
\begin{tabular}{l|cccc}
\multirow{2}{*}{pretrain} & \multicolumn{4}{c}{\# of training sample} \\
& 50 & 100 & 150 & 200\\\specialrule{.15em}{.05em}{.05em}
\textcolor{gray}{scratch} & 89.5~~~~~~~~~~~ & 91.7~~~~~~~~~~~ & 93.0~~~~~~~~~~~ & 94.1~~~~~~~~~~~ \\
self-sup. & 92.3 \textcolor{teal}{(+2.8)} & 94.3 \textcolor{teal}{(+2.6)} & 95.8 \textcolor{teal}{(+2.8)} & 96.4 \textcolor{teal}{(+2.3)} \\
\end{tabular}
}
\vspace{-0.3cm}
\caption{{\bf OA for various number of training samples.}}
\label{tab:num_of_training_samples}
\end{table}

\noindent{\bf \# of training samples.} We compare our model with the model trained from scratch by varying the number of training samples, as shown in Table~\ref{tab:num_of_training_samples}. As expected, our model using self-supervised learning yields improved accuracy over its counterpart model, when using any number of training samples. It is noteworthy that the OA gap was larger when using a small number of training samples. This observation indicates that training the model from scratch suffers more from overfitting due to the smaller number of training samples, but using a pre-trained model mitigates this overfitting well.

\subsection{Backbone Modification}
\label{ssec:backbone_modification}

\begin{table}[t]
\captionsetup{font=small}
\setlength{\tabcolsep}{6.0pt}
\renewcommand{\arraystretch}{0.9}
\centering
{\small
\begin{tabular}{cc|cc}
no multi-scale & more res. module & OA & AA \\\specialrule{.15em}{.05em}{.05em}
$\cdot$ & $\cdot$ & 92.0~~~~~~~~~~~ & 91.2~~~~~~~~~~~ \\
\checkmark & $\cdot$ & 92.7 \textcolor{teal}{(+0.7)} & 91.9 \textcolor{teal}{(+0.7)} \\ 
$\cdot$ & \checkmark & 92.9 \textcolor{teal}{(+0.9)} & 92.0 \textcolor{teal}{(+0.8)} \\
\checkmark & \checkmark & {\bf 94.1} {\bf \textcolor{teal}{(+2.1)}} & {\bf 92.3} {\bf \textcolor{teal}{(+1.1)}} \\
\end{tabular}
}
\vspace{-0.3cm}
\caption{{\bf Backbone modification.} In the parentheses, gaps to the original backbone~\cite{HLeeTIP2017} are shown.}
\label{tab:backbone}
\end{table}

The backbone used in the cross-domain CNN~\cite{HLeeTIP2017} that our model inherits has been modified with two factors: i) the initial multi-scale layer is replaced by a single $5\times 5$ layer (``no multi-scale''), and ii) all remaining layers take a form of residual modules (``more res. module''). Each modification factor is compared in terms of accuracy in Table~\ref{tab:backbone}. Both individual modifications can improve the accuracy and the new backbone employing all these modifications presents the best accuracy. Due to the superior performance of the modified backbone, this backbone has been used in all the experiments.

\section{Conclusion}
\label{sec:conclusion}

The proposed self-supervised learning method showed promising results compared to its supervised counterpart. The improved accuracy has been derived from enriching the feature learning in a common representation space using multiple unlabeled hyperspectral images with different spectral properties. We also carried out a comprehensive study to search for the optimal parameters (i.e., the sample region size $p$, \# of training samples, and backbone modifications). We believe that the proposed work can potentially advance the learning of multi-dimensional remote sensing data as our method can incorporate diverse hyperspectral images without annotated labels while providing advanced classification performance.

\bibliographystyle{IEEEbib}
\bibliography{refs}

\end{document}